%% file: _main.tex
\input{_constants}
\arxiv %

\pdfoutput=1
\documentclass[10pt,twocolumn,letterpaper]{article}
\input{cvpr_header}

\unless\ifarxiv \myexternaldocument{_supplementary} \fi

\makeatletter
\renewcommand\AB@affilsepx{ \hphantom{------} \protect\Affilfont}
\makeatother

\makeatletter
\def\@maketitle{%
  \newpage
  \null
  \vskip 2em%
  {\begin{center}%
   \let \footnote \thanks
   {\Large\bfseries \@title \par}%
   \vskip 2em%
   {\large \AB@authlist\par}%
   \vskip 1em%
   {\large \AB@affillist\par}%
  \end{center}}%
  \par
  \vskip 1.5em}
\makeatother

\author[1,2]{Alex Trevithick}
\author[2]{Roni Paiss}
\author[3]{Philipp Henzler}
\author[2]{Dor Verbin}
\author[2,4]{Rundi Wu}
\author[2,5]{Hadi Alzayer}
\author[2]{Ruiqi Gao}
\author[2]{Ben Poole}
\author[2]{Jonathan T. Barron}
\author[2]{Aleksander Holynski}
\author[1]{Ravi Ramamoorthi}
\author[2]{Pratul P. Srinivasan}

\affil[1]{UC San Diego} 
\affil[2]{Google DeepMind}
\affil[3]{Google Research}
\affil[4]{Columbia}
\affil[5]{Univ. Maryland}

\begin{document}
\title{\paperTitle}

\twocolumn[{
\renewcommand\twocolumn[1][]{#1}
\maketitle

\begin{center}
    \includegraphics[width=0.95\linewidth]{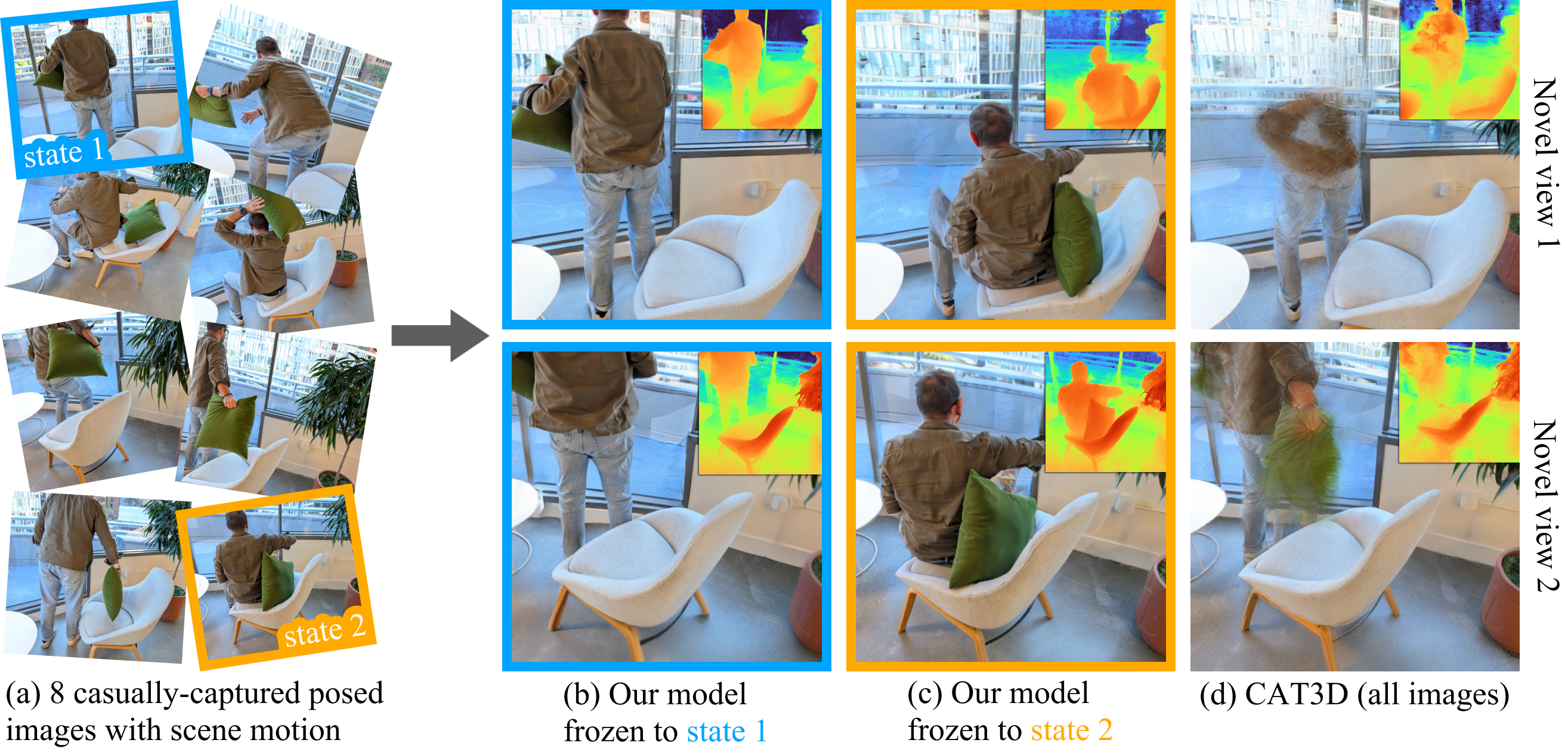}
\captionof{figure}{We show results of our model applied to a casual in-the-wild capture. (a) Given $8$ unordered images of a scene with significant motion and desired states marked in blue and orange, our model generates a 3D representation for each desired state shown in corresponding colors in (b) and (c). The CAT3D baseline~\cite{cat3d} in (d) cannot disentangle the different states, resulting in catastrophic failure.}
\label{fig:teaser}
\end{center}
}]

\input{00_abstract}

\input{01_intro}
\input{02_related}

\input{03_method}

\input{05_results}

\input{10_conclusion}

{\small
\bibliographystyle{ieeenat_fullname}
\bibliography{11_references}
}


\end{document}


\title{\paperTitle}
\author{\authorBlock}
\maketitlesupplementary

\appendix
\input{12_appendix}

{\small
\bibliographystyle{ieeenat_fullname}
\bibliography{11_references}
}

%% file: _constants.tex
\def\paperID{2214}
\def\confName{CVPR}
\def\confYear{2025}

\newif\ifreview 
\newif\ifarxiv \newcommand{\arxiv}{\arxivtrue}
\newif\ifcamera 
\newif\ifrebuttal 

%% file: cvpr_header.tex
\ifreview \usepackage[review]{cvpr} \fi
\ifarxiv \usepackage[pagenumbers]{cvpr} \fi
\ifrebuttal \usepackage[rebuttal]{cvpr} \fi
\ifcamera \usepackage{cvpr} \fi

\input{_macros}  %

\usepackage{xr-hyper}

\makeatletter
\newcommand*{\addFileDependency}[1]{
  \typeout{(#1)}
  \@addtofilelist{#1}
  \IfFileExists{#1}{}{\typeout{No file #1.}}
}

\makeatother
\newcommand*{\myexternaldocument}[1]{
    \externaldocument{#1}
    \addFileDependency{#1.tex}
    \addFileDependency{#1.aux}
}

\definecolor{cvprblue}{rgb}{0.21,0.49,0.74}
\usepackage[pagebackref,breaklinks,colorlinks,citecolor=cvprblue]{hyperref}
\usepackage[capitalize]{cleveref}
\crefname{section}{Sec.}{Secs.}
\crefname{table}{Table}{Tables}
\crefname{figure}{Fig.}{Figs.}

\ifarxiv \crefname{appendix}{App.}{Apps.}
\else \crefname{appendix}{Suppl.}{Suppls.} \fi

%% file: _macros.tex
\usepackage{graphicx}	

\usepackage{amsmath}	
\usepackage{amssymb}	
\usepackage{booktabs}
\usepackage{times}
\usepackage{microtype}
\usepackage{epsfig}
\usepackage{caption}
\usepackage{float}
\usepackage{placeins}
\usepackage{color, colortbl}
\usepackage{stfloats}
\usepackage{enumitem}
\usepackage{tabularx}
\usepackage{xstring}
\usepackage{multirow}
\usepackage{xspace}
\usepackage{url}
\usepackage{subcaption}
\usepackage[hang,flushmargin]{footmisc}
\usepackage{subcaption}
\usepackage{makecell} %
\usepackage{multirow} %
\usepackage{arydshln} %
\makeatletter
\robustify\@latex@warning@no@line
\makeatother
\usepackage{authblk}

\newcommand{\name}{SimVS}

\def\paperTitle{\name: Simulating World Inconsistencies for Robust View Synthesis}

\ifcamera \usepackage[accsupp]{axessibility} \fi

\ifarxiv  \fi

\newcommand{\R}[1]{{%
    \textbf{%
        \ifstrequal{#1}{1}{\textcolor{red}{R#1}}{%
        \ifstrequal{#1}{2}{\textcolor{blue}{R#1}}{%
        \ifstrequal{#1}{3}{\textcolor{magenta}{R#1}}{%
        \ifstrequal{#1}{4}{\textcolor{teal}{R#1}}{%
                           \textcolor{cyan}{R#1}%
        }}}}%
    }%
}}

\newcommand\lft{\mathopen{}\left}
\newcommand\rgt{\aftergroup\mathclose\aftergroup{\aftergroup}\right}

\newcommand{\myparagraph}[1]{ \vspace{3pt}  \noindent {\bf #1}\,\,\,}

%% file: 00_abstract.tex
\begin{abstract}
Novel-view synthesis techniques achieve impressive results for static scenes but struggle when faced with the inconsistencies inherent to casual capture settings: varying illumination, scene motion, and other unintended effects that are difficult to model explicitly. We present an approach for leveraging generative video models to simulate the inconsistencies in the world that can occur during capture. We use this process, along with existing multi-view datasets, to create synthetic data for training a multi-view harmonization network that is able to reconcile inconsistent observations into a consistent 3D scene. We demonstrate that our world-simulation strategy significantly outperforms traditional augmentation methods in handling real-world scene variations, thereby enabling highly accurate static 3D reconstructions in the presence of a variety of challenging inconsistencies.

\end{abstract}

%% file: 01_intro.tex
\section{Introduction}
\label{sec:intro}

View synthesis, the task of creating images from unobserved camera viewpoints given a set of posed images, has seen remarkable progress in recent years. Current algorithms are able to render detailed photorealistic novel views of complicated 3D scenes. However, these techniques tend to assume that the provided input images are \emph{consistent} --- that the geometry and illumination of the scene is static during capture. Typical captures of real-world scenes seldom obey this constraint; people and objects may move and deform, and lights may move or change brightness.

Moreover, casual captures outside of tightly-controlled settings tend not only to be inconsistent but also \emph{sparse}, containing only a small number of observed views. Methods for sparse view synthesis are usually trained on synthetic or captured multiview datasets that are consistent by design, and therefore fail to generalize to the inconsistencies seen in real-world casual captures (see \cref{fig:cat3d_bad} as an example).

We address the problem of robust view synthesis from sparse captures in a new way by leveraging the ability of video diffusion models to simulate plausible world inconsistencies that could arise during capture. 
There has been considerable speculation about the usefulness of large video models as simulators or ``world models''~\cite{sora, veo2024} and in this work we demonstrate a new use case for their simulation capabilities.

Our approach augments existing multiview datasets with inconsistencies simulated by a video diffusion model and trains a multiview harmonization model to sample sets of consistent views of a scene conditioned on sparse inconsistent captures. We can then use existing 3D reconstruction and view synthesis techniques to synthesize novel viewpoints from these consistent images.
\\ \indent
In summary, our key technical contributions are:
\begin{itemize}
\item A generative data augmentation strategy that leverages video diffusion models to sample world inconsistencies (\eg scene motion and lighting changes) that could arise during capture (Section~\ref{sec:vidmodel})
\item A multiview harmonization model, trained on this generated data, that converts inconsistent sparse input images into a set of consistent images (Section~\ref{sec:harmonization})
\end{itemize}
We demonstrate that our \emph{generative augmentation} strategy outperforms other alternatives such as using heuristic data augmentation or synthetic rendered data, and that novel views rendered from our harmonization model are superior to those from existing approaches for sparse and robust view synthesis. We encourage readers to view our video results in the supplement.

%% file: 02_related.tex
\begin{figure}
    \centering
\includegraphics[width=\columnwidth]{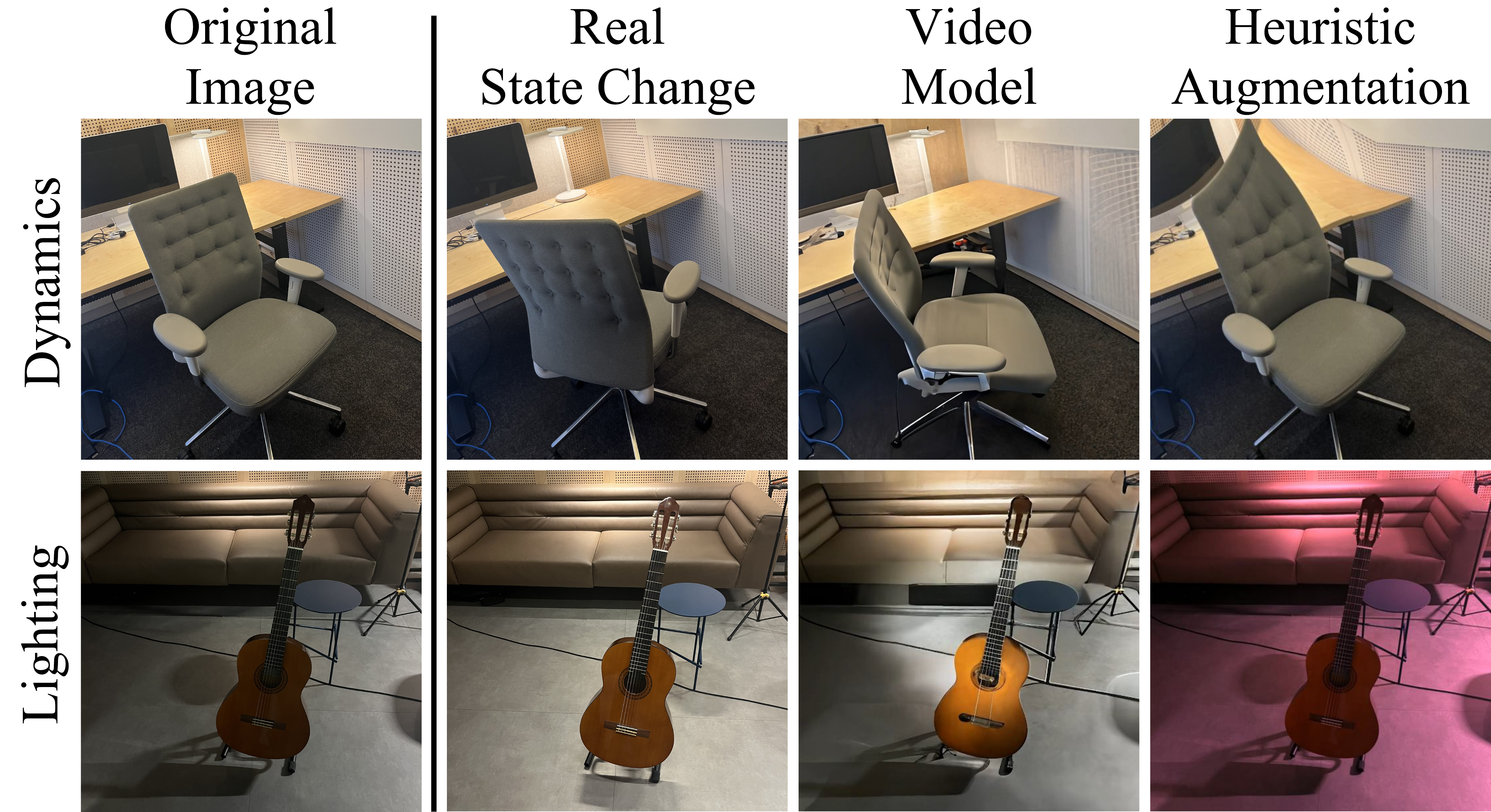}
    \caption{A comparison of real world state changes, those simulated through a video model, and heuristic augmentations (random sparse flow fields for dynamics and random color tints for lighting).} %
    \label{fig:data}
\end{figure}

\section{Related Work}
We address the task of view synthesis from sparse \emph{and} inconsistent images of a scene. While existing techniques address view synthesis from densely-sampled inconsistent inputs or sparse consistent inputs, to our knowledge no existing method is capable of synthesizing novel views of full scenes from images that are both sparse and inconsistent.

\subsection{Robust view synthesis}
Prior methods for robust view synthesis typically require dense captures with hundreds of images and focus on explicitly modeling a specific source of inconsistency (either motion or lighting) as part of reconstruction.

\paragraph{Scene dynamics}
In the case of scene dynamics, existing methods start with a dense video and attempt to recover motion flows or trajectories to explain the observed motion. 
Early approaches based on Neural Radiance Fields~\cite{mildenhall2020nerf} optimized time-varying flow fields to explain observed motion as deformations of an underlying consistent scene representation~\cite{park2021nerfies,nsff,dnerf,nrnerf,hypernerf,tineuvox}. Later NeRF-based methods improved quality further through prior integration~\cite{rodynrf,li2023dynibar}. The most recent state-of-the-art methods have adopted 3D Gaussian representations and optimized explicit motion trajectories for this particle-based scene representation~\cite{lei2024mosca,gaussianmarbles,som2024}, often leveraging strong priors from pretrained monocular depth~\cite{marigold, geowizard, depth_anythingv2}, optical flow~\cite{raft}, or tracking models~\cite{tapir, cotracker, spatialtracker}. 
These temporal priors have been crucial for rendering high-quality novel views in the dense capture setting, but tend to break down when applied to sparse or unordered captures.

\paragraph{Lighting inconsistencies}
Existing structure-from-motion pipelines display remarkable robustness to lighting variation~\cite{colmap, superglue, hloc}, enabling 3D reconstruction from large-scale in-the-wild images~\cite{buildingrome}. 
To model inconsistencies due to changing scene lighting, 3D reconstruction and view synthesis techniques use per-image ``appearance embeddings'' that allow for the appearance of scene content to vary across observations~\cite{meshry2019neural,martinbrualla2020nerfw,kulhanek2024wildgaussians,gaussianwild,swag,wegs,wildgs}. This strategy can successfully model lighting inconsistencies given dense captures with smoothly-varying appearance changes, but is unable to reconcile large changes in appearance in sparsely-sampled captures.

\subsection{Sparse view synthesis}
In novel view synthesis settings with only a few captured views, most methods rely on strong priors learned from large multiview datasets.
Some methods train feedforward models to directly predict 3D representations that can be used for view synthesis~\cite{grf, pixelnerf, mvsnerf, geonerf, lrm, longlrm}. Others rely on pretrained monocular depth, multiview stereo, or inpainting networks and rely on test-time optimization to fit a scene~\cite{instantsplat, realmdreamer, sparf, shih2024extranerf, weber2024nerfiller}. 
A recent class of methods has achieved high visual quality by directly generating images from novel viewpoints using diffusion models conditioned on observed image(s) and target camera poses~\cite{zero123,12345, long2023wonder3d, shi2023MVDream, cat3d, wu2024reconfusion}. In particular, the multiview diffusion model CAT3D~\cite{cat3d} has emerged as the state-of-the-art in view synthesis from sparse image inputs. However, as they are trained only on consistent multiview image sets, CAT3D and other sparse view synthesis techniques are not robust to the inconsistencies observed in real-world casual captures. Concurrent work CAT4D~\cite{cat4d} extends CAT3D with a temporal axis. Among other training data, it leverages the generative augmentation strategy proposed in this paper.

\label{sec:related}

%% file: 03_method.tex
\begin{figure*}
    \centering
    \includegraphics[width=\textwidth]{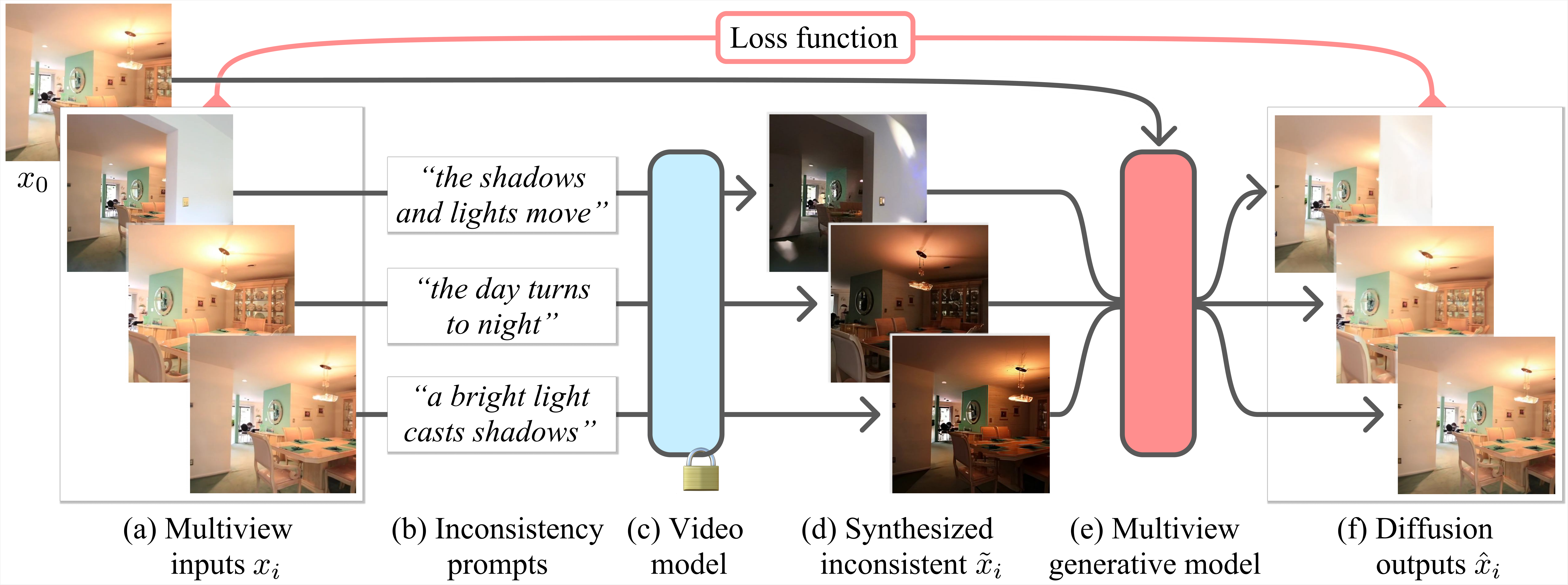}
    \caption{Our method's overall pipeline. (a) Given a dataset of multiview images $x_i$, we simulate inconsistencies by (b) prompting a (c) video model and then (d) selecting inconsistent frames $\Tilde{x}_i$. We feed these images along with a held-out reference image $x_0$ under the original condition to a (e) multiview generative model to predict (f) a set of corresponding consistent outputs $\hat x_i$. This output is supervised by the original multiview images $x_i$.}
    \label{fig:pipeline}
\end{figure*}

\section{Simulating World Inconsistencies with Video Models}
\label{sec:vidmodel}
Training a robust view synthesis model is challenging due to the lack of paired training data of inconsistent captures and target consistent images. Most existing multiview datasets only contain captures of \emph{consistent} scenes, so simply scaling such data is not sufficient for robust view synthesis. Gathering images from multiple viewpoints, each under multiple scene deformations or lighting settings, would be extremely onerous. Heuristic data augmentation strategies such as random transformations, tints, and sparse flow fields cannot adequately capture the diversity and 3D nature of scene motions and lighting changes, as displayed in~\cref{fig:data}. Conversely, synthetic datasets like Objaverse~\cite{Deitke2022ObjaverseAU} only contain simple object-level motion and fail to enable generalization to real-world scenes. %

The key idea in our work is to leverage generative video models to create a robust view synthesis dataset from existing consistent multiview image datasets. For each 3D scene, we desire a dataset that contains (1) a set of consistent multiview images $x_i$, (2) inconsistent images $\Tilde{x}_i$ where the scene has undergone some transformation such as a deformation or lighting change, and (3) camera poses $\pi_i$ for each image.

\subsection{Video model augmentation}

We propose to generate a realistic and diverse dataset of inconsistent conditioning images by simulating dynamic motion and lighting inconsistencies with pretrained image-to-video generative models. Starting with a multiview capture (taken from existing large-scale multiview image datasets), we first generate, for each view, a video from a static camera with simulated scene changes (motion or lighting). By sampling frames from these videos, we can obtain inconsistent observations for each captured viewpoint. Other image editing approaches such as InstructPix2Pix~\cite{ip2p} could potentially be used to perform this inconsistency transformation, but these methods often fail to produce substantial variation in the layout of the image which are needed to simulate dynamic inconsistency.

To generate videos with simulated scene changes, we use an image- and text-conditioned video diffusion model that samples from $p(v \vert I, c)$, where $V$ is a video, $I$ is a conditioning frame that $V$ should include, and $c$ is a text caption. By setting $I$ to an image from a multiview capture $x_i$ and choosing $c$, we may simulate inconsistencies on top of the image. Note that the video must not contain camera motion in order to preserve the accuracy of existing camera parameters. 

We simulate the two most prominent inconsistencies: dynamic motion and lighting changes.
For dynamics, we use the Mannequin Challenge dataset~\cite{mannequinchallenge}. This dataset is a natural choice as it includes static multiview captures of scenes with content that would typically be dynamic in casual captures.
For our lighting-robust model, we simulate lighting changes on the RealEstate10k~\cite{zhou2018stereo} dataset, which contains scenes in diverse indoor and outdoor illumination conditions.

We generate the captions $c$ with a multimodal large language model, Gemini~\cite{team2023gemini}.  For each clip in the dataset, we randomly choose a representative frame $x_i$. We prompt Gemini with this frame and a meta-prompt $m$, designed to elicit simple but specific prompts, e.g., ``the woman swings the pillow'' or ``the two children dance.'' We also ensure the generated prompts are sufficiently specific and concise through $m$ (see the supplement for the meta-prompt in its entirety). We generate the complete inconsistency prompt as:
\begin{equation}
    c = \text{``static shot, ''} \oplus \operatorname{Gemini}(x_i, m)\,,
\end{equation}
where $\oplus$ denotes string concatenation. Since the inconsistencies observed in casual captures of dynamic content are highly correlated across views, we use the same inconsistency prompt for all frames in the corresponding clip.

We find that incorporating a negative prompt~\cite{negativeprompt} $c_\text{negative}$ is extremely important for generating the desired inconsistencies without changing the camera viewpoint. We include phrases such as ``panning view'' and ''orbit shot`` in $c_\text{negative}$.

Given a multiview image $x_i$ and generated inconsistency prompt $c$, we sample a video:
\begin{equation}
    V = \operatorname{VideoModel}\lft(x_i, c, c_\text{negative}\rgt)\,.
\end{equation}
We assume all frames in the sampled video are inconsistent with respect to the original image $x_i$. Therefore, we get a set of $T$ inconsistent frames corresponding to $x_i$ for each sample from the video model, where $T$ is the number of frames in the output video. At training time, we randomly sample one of the $T$ frames as inconsistent conditioning for a given $x_i$. 
In our experiments, we generate $640$ total ``inconsistent'' frames per multiview capture, giving us a total of about $6$ million frames for the dynamics dataset and about $12$ million frames for the lighting dataset. Figs.~\ref{fig:data} and \ref{fig:pipeline} visualize example frames from our synthesized videos. 

\begin{figure}
    \centering
    \includegraphics[width=\columnwidth]{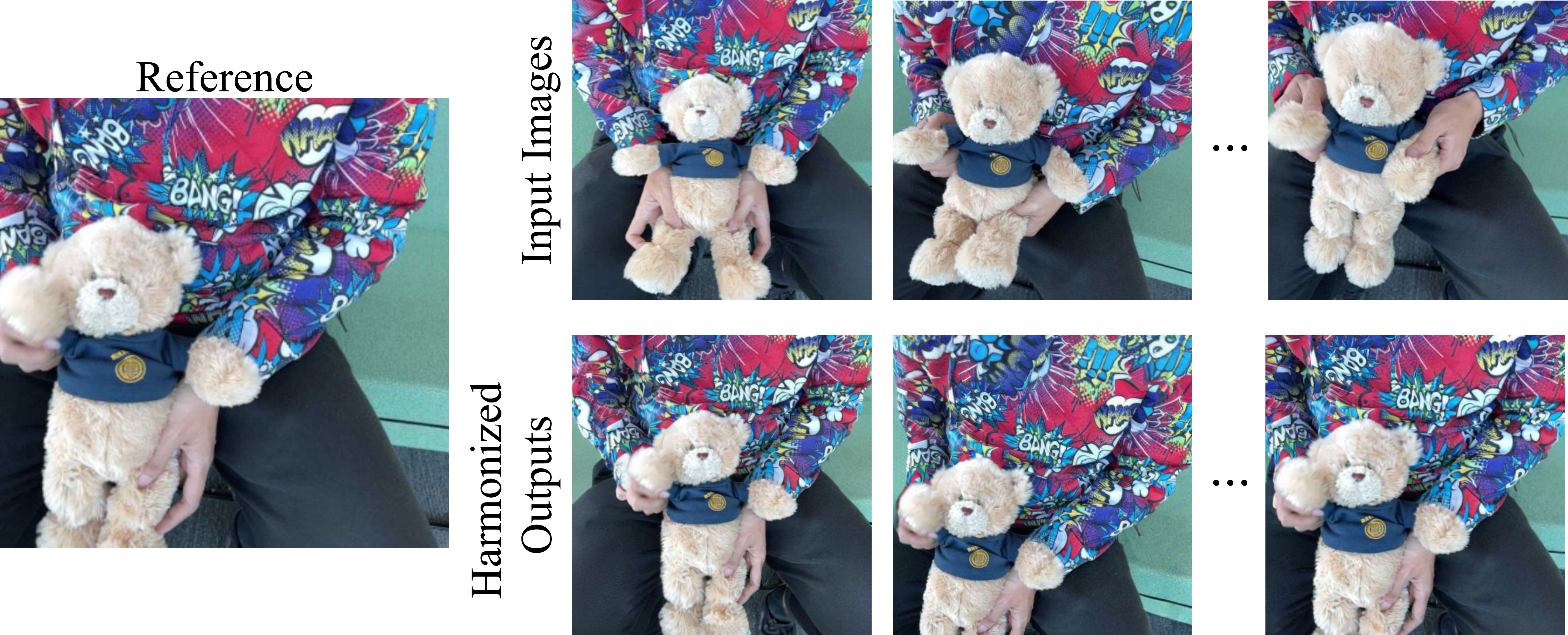}
    \caption{Samples from our multiview diffusion harmonization model, visualized for scene dynamics. Given the reference image and inconsistent input image, our model directly generates multiview images consistent with the state of the reference.}
    \label{fig:harmonization}
\end{figure}

\begin{figure}
    \centering
    \includegraphics[width=\columnwidth]{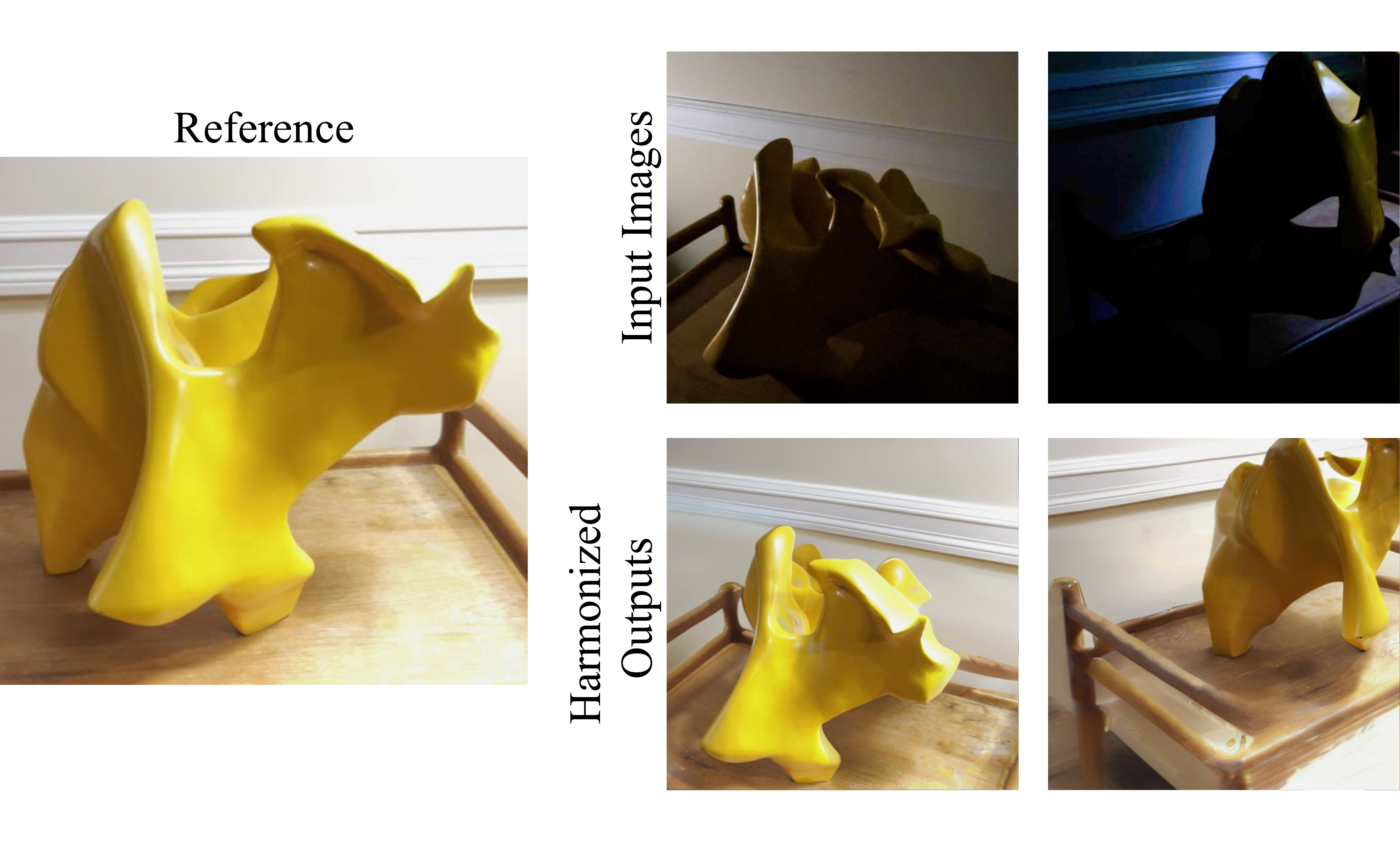}
    \caption{Samples from our multiview diffusion harmonization model, visualized for lighting. Given the reference image and inconsistent input image, our model directly generates multiview images consistent with the state of the reference.}
    \label{fig:light_harmonization}
\end{figure}

\subsection{Video model details}
For all experiments in this paper, we use Lumiere~\cite{bar2024lumiere}, a pixel-space video diffusion model which operates in two-stages for high-resolution generation. We find that the Lumiere model struggles to generate lighting changes when given non-generic prompts, so for our lighting-robust model, we sample $c$ uniformly from a set of predetermined lighting prompts found to generate large lighting variations instead of using a large language model. Please refer to~\cref{fig:pipeline} and the supplement for example prompts.

\begin{figure*}[h!]
    \centering
    \includegraphics[width=\textwidth]{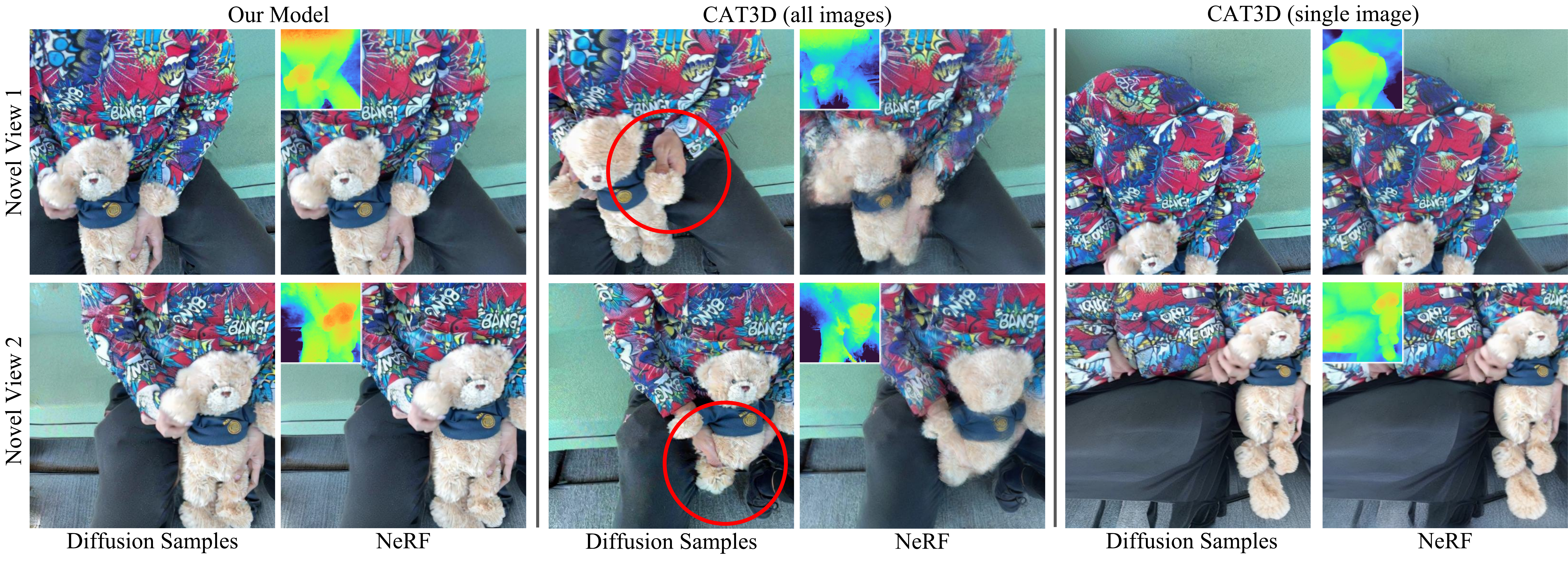}
    \vspace{-0.2in}
    \caption{Given the reference and inputs in~\cref{fig:harmonization}, we show the outputs of our model versus the CAT3D baselines. We display both the diffusion samples and learned NeRF representations with the depth maps inset. Note the 3D consistency of our samples in comparison to the changing articulation of the scene displayed by CAT3D taking all images as input. The single-image conditional CAT3D has no notion of scene scale, cannot use multiview cues to reason about the static parts, and must hallucinate all scene content outside of the input image's frustum. }
    \label{fig:cat3d_bad}
\end{figure*}

\section{Harmonization through multiview diffusion}
\label{sec:harmonization}
We use our multiview simulated world inconsistencies dataset $(x, \Tilde{x}, \pi)$ to learn a generative model that can map from sparse inconsistent captures to a consistent set of images, as displayed in Figs.~\ref{fig:harmonization} and~\ref{fig:light_harmonization}. We call this model a ``harmonization'' model as it brings the inconsistent input images into harmony.

\subsection{Architecture}\label{subsec:model}
We build our harmonization model on top of CAT3D~\cite{cat3d}, a latent multiview diffusion model that directly predicts target images conditioned on posed input images and target camera poses. 
To incorporate inconsistent observed images as conditioning, we simply concatenate latents of the inconsistent images $\Tilde{z}_i=\mathcal{E}(\Tilde{x}_i)$, encoded by the VAE encoder $\mathcal{E}$, to the target raymaps and noisy latents. Additionally, we concatenate a binary image mask (either all ones or all zeros) to each input to denote the reference image, i.e., the ``desired state'' with which all other outputs should be consistent.

\subsection{Training}\label{subsec:train}
Our goal is to learn a generative model that produces consistent output image sets with $N$ images, given a reference image latent $z_0$ signifying the desired scene state and $n\leq N$ observed inconsistent image latents $\Tilde{z}_i$:
\begin{equation}
p\lft(z_{1:N} \mid z_0, \Tilde{z}_{1:n}, \pi_{0:N}\rgt)\,. \label{eq:sample}
\end{equation}
Given a reference conditioning latent $\mathcal{E}(x_0)=z_0$ and up to $7$ inconsistent posed latents $\Tilde{z_i}$, our model predicts latents $z_{1:7}$ corresponding to the ground-truth consistent image latents. We finetune our model parameters from CAT3D~\cite{cat3d}, with additional parameters in the first layer to account for the additional conditioning channels. We train the model with the same diffusion loss and weighting as \cite{cat3d}:
\begin{align}
\label{eq:welovemath}
\!\!\mathbb{E}_{t,\epsilon,z_{0:7}, \tilde{z}_{1:7}}\lft[w(t)\lft\|f\lft(\alpha_t z_{1:7} + \sigma_t\epsilon; z_0, \tilde{z}_{1:7}\rgt) - z_{1:7}\rgt\|^2\rgt]\,,
\end{align}
where $f$ is our multiview diffusion ``harmonization'' model. Given noisy versions of the target consistent latents $z_{1:7}$, the target conditioning latent $z_0$, and the inconsistent inputs $\tilde{z}$, we aim to produce a denoised estimate output by $f$ that is as close as possible to the consistent latents $z_{1:7}$. We additionally uniformly drop out the number of conditioning frames $\Tilde{z}$ to allow the model to handle between $1$ and $8$ input images at test time.

\subsection{3D reconstruction}
Having trained the harmonization model, we can sample consistent latents $\hat z_{1:7}$ and decode them into images $\hat{x}_{1:7}$ with the VAE decoder (visualized in Figs. \ref{fig:harmonization}, \ref{fig:light_harmonization}, and \ref{fig:cat3d_bad}). We then have a total of 8 consistent images: the initial observed target $x_0$ and model outputs $\hat x_{1:7}$. While 3D reconstruction from such a small image collection is infeasible, we can use multiview diffusion models trained on consistent images such as CAT3D~\cite{cat3d} to ``densify'' the sparse consistent capture into a dense consistent capture with enough views to train a NeRF. Instead of directly sampling the original $3$-image conditional CAT3D model, we finetune it to condition on $5$-frames, finding that the additional context outperforms the original $3$-image conditional model in our setting.

\begin{figure*}
    \centering
    \includegraphics[width=\textwidth]{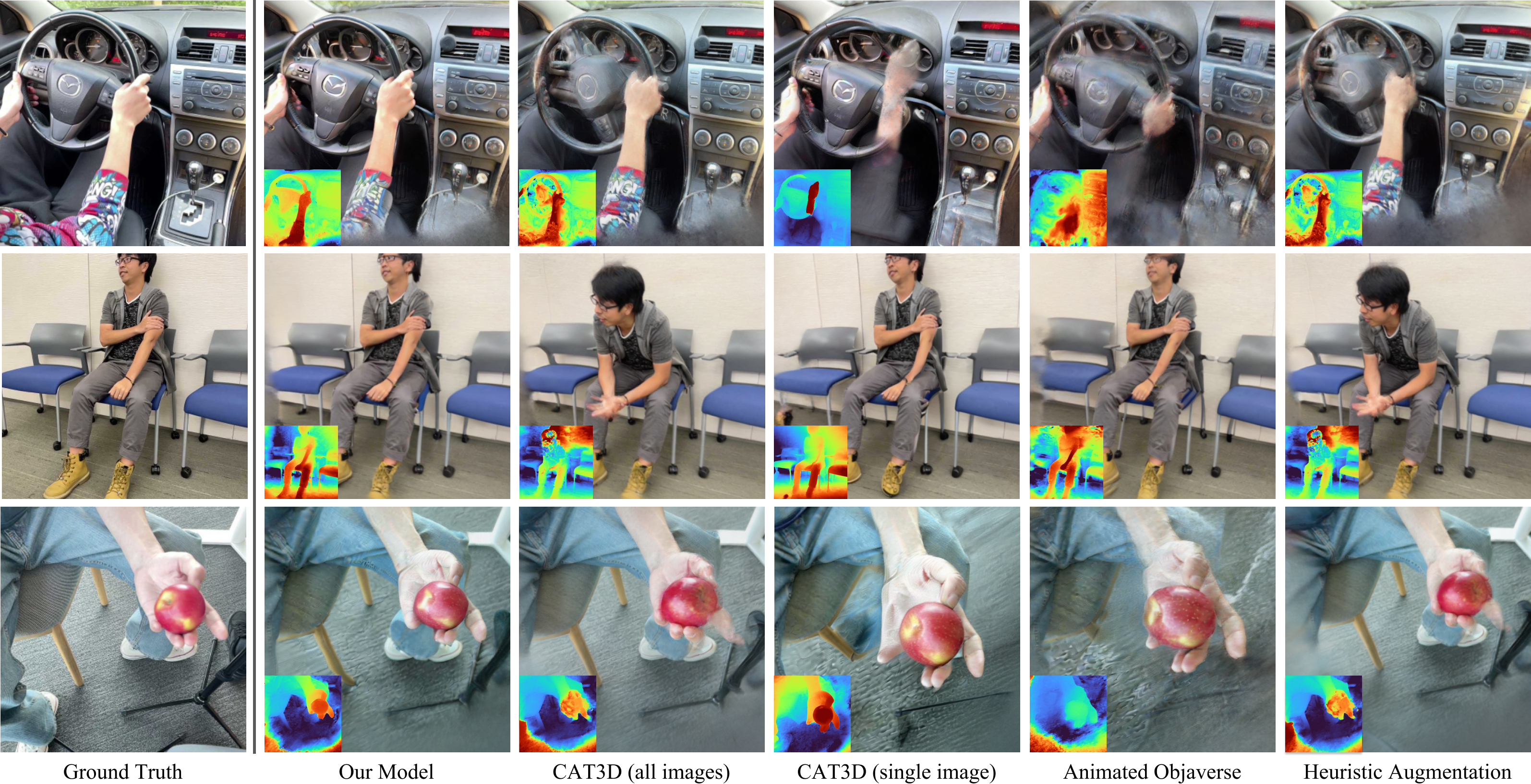}
    \caption{Qualitative results for the DyCheck~\cite{dycheck} dataset for our model, two CAT3D baselines, and two of our ablations. The depth maps are inset on the bottom left. Images are cropped for visualization. Compared to CAT3D (all images), our method generates coherent 3D scenes despite the scene motion, while leveraging the information from multiple input views unlike CAT3D (single image). In comparison to the ablations, the quality of our approach is superior.}
    \label{fig:dynamics_qual}
\end{figure*}

\begin{figure*}
    \centering
\includegraphics[width=\textwidth]{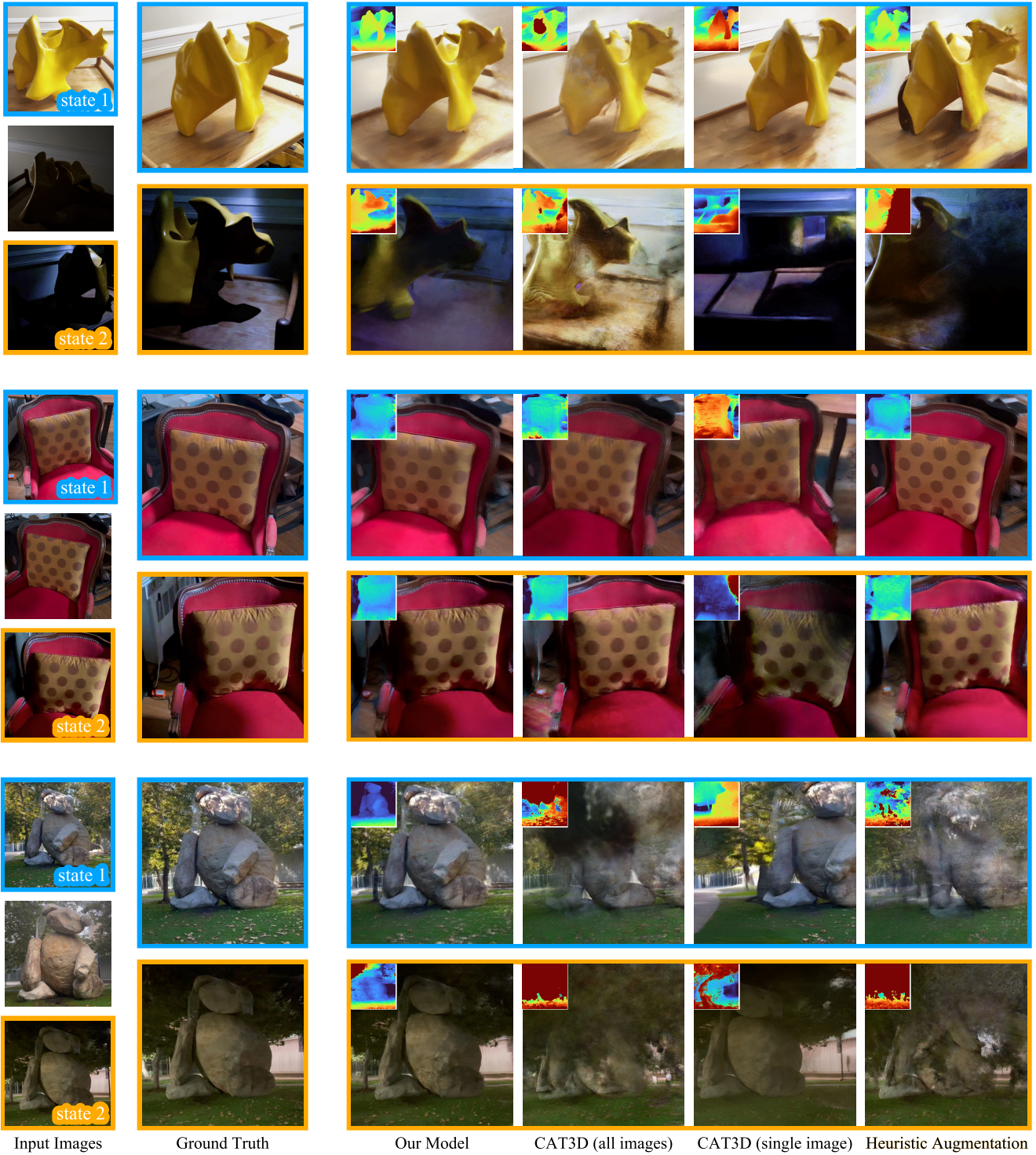}
    \caption{Qualitative results for $3$ scenes from our captured lighting dataset. For each scene, we display the renders from the learned NeRFs given the $3$ input images on the left. We show two states for each scene, with the renders outlined in blue corresponding to the upper input image, and the renders outlined in orange corresponding to the bottom input image. Although all methods are rendered with the appearance embedding of the corresponding states on the left, baselines struggle to generate plausible novel views, and often generate completely degenerate geometry. The bottom rows are brightened for visualization. }
    \label{fig:lighting_qual}
\end{figure*}

%% file: 05_results.tex
\section{Experiments}
We evaluate our method for the two most common sources of inconsistency during casual multiview capture: scene dynamics and lighting changes. 

\subsection{Scene Dynamics}

\begin{table}
\centering
\begin{tabular}{@{}l|ccc@{}}
        Method   & PSNR$\uparrow$ & SSIM$\uparrow$ & LPIPS$\downarrow$ \\ \midrule 
CAT3D (single image)  &   14.61   &             0.382            &        0.473                                       \\

CAT3D (all images) &   15.59   &             0.448            &        0.462                                       \\ \midrule 

Our Model        &      \textbf{16.73}    &            \textbf{0.463}            &         \textbf{0.413}                                           \\
\end{tabular}
\caption{View synthesis results on the DyCheck dataset~\cite{dycheck} comparing our model to CAT3D taking one or all conditioning images. Our model outperforms CAT3D by all metrics.}
\label{tab:dynamic_comparisons}
\end{table}

\myparagraph{Dataset}
For scene dynamics, we evaluate our method on DyCheck~\cite{dycheck}, a dataset of $7$ multiview videos where the assumed input is a monocular video with significant scene and camera motion. In this setting, we select $7$ sparse frames uniformly in time as a consistent conditioning set, and uniformly select $4$ target time images (top left of \cref{fig:harmonization}) per scene for which to compute metrics. Note that prior works which handle scene dynamics assume an ordered dense capture~\cite{lei2024mosca, som2024,rodynrf}. 

\myparagraph{Baselines}
Considering this task as view synthesis from $8$ inputs, we compare our performance to the state-of-the-art method for sparse view synthesis, CAT3D~\cite{cat3d}. We evaluate variants of CAT3D which take all of the images as input CAT3D (all images), and only one of the images as input CAT3D (single image). For CAT3D (all images), we find that a finetuned model which conditions on $5$ images and predicts $3$ instead of conditioning on $3$ and predicting $5$ works slightly better for this setting. When sampling target views, we always include the reference image in the conditioning set, along with the $4$ closest of the $7$ views to the current target camera set. CAT3D (single image) simply receives only the ground truth reference image. 

Due to noisy camera poses and the underdetermined nature of our task, we recompute the poses per method using COLMAP on the samples and train a Zip-NeRF~\cite{barron2023zipnerf} to evaluate novel view synthesis quality. In $4$ of the $28$ timesteps, COLMAP was unable to register the test images for at least one of the baselines; we discard those scenes from the calculation. Note that COLMAP never fails to register the test images for our method's results.  

\myparagraph{Results}
The quantitative results shown in~\cref{tab:dynamic_comparisons} demonstrate that we significantly outperform CAT3D across all metrics. Qualitatively, we can see in Figs.~\ref{fig:teaser},~\ref{fig:cat3d_bad} and~\ref{fig:dynamics_qual} that CAT3D simply cannot handle inconsistencies. Their diffusion samples display high variance, typically changing state based on proximity to the input views. Training a 3D representation such as NeRF from these samples leads to undesirable averaging over all of the states and significant blur in inconsistent regions. Please see our supplement for video comparisons.

\begin{table}
\centering
\begin{tabular}{@{}l|ccc@{}}
        Method   & PSNR$\uparrow$ & SSIM$\uparrow$ & LPIPS$\downarrow$ \\ \midrule 
CAT3D (single image)          &       15.06                  &          0.526             & 0.552                            \\
CAT3D (all images) &   18.26   &             0.625            &        0.419                                        \\  \midrule 
Our Model        &     \textbf{20.98}      &            \textbf{0.707}     &     \textbf{0.357}                                     \\
\end{tabular}
\caption{View synthesis results on our lighting variation dataset comparing our model to CAT3D taking one or all images as input. Our model outperforms the baselines with the appearance embedding of the target image.}
\label{tab:lighting_comparisons}
\end{table}

\subsection{Lighting Changes}

\myparagraph{Dataset}
For lighting changes, we are not aware of an existing dataset of posed images that contains multiple illumination conditions and multiple ``ground truth'' images under a consistent lighting. Note that the widely-used Phototourism dataset~\cite{phototourism} contains only one image under each lighting. We create a new dataset of real-world scenes captured under $3$ separate lighting conditions. To construct this dataset, we take $3$ monocular videos of a scene in $3$ different lighting conditions, using approximately the same camera trajectory for each. 

Using $3$ frames (one from each inconsistent video) as input, we evaluate renderings of the held-out images from one of the lighting conditions. For each collected scene, we select a target illumination condition and evaluate the method's abilities to do novel-view synthesis for that illumination given the three images. See~\cref{fig:lighting_qual} for example inputs and targets. We use Hierarchical Localization~\cite{hloc} with SuperGlue~\cite{superglue} feature matching to jointly pose all images.

\myparagraph{Baselines}
Methods which handle lighting changes typically assume a large number of captured images~\cite{martinbrualla2020nerfw, kulhanek2024wildgaussians} and rely on latent embeddings~\cite{bojanowski2018optimizing} to parameterize per-image variations in appearance. To create a strong baseline, we first generate a large number of novel views using CAT3D conditioned on the 3 inconsistent images, and then train a Zip-NeRF with latent appearance embeddings. At test-time, for all methods, we use the embedding of the reference image with the target illumination. We again evaluate against CAT3D (all images) and CAT3D (single image). However, since there are only $3$ input images, CAT3D (all images) is the original CAT3D model conditioned on the three input images. 

\myparagraph{Results}
The quantitative results in~\cref{tab:lighting_comparisons} show that our method significantly outperforms CAT3D in all metrics. Qualitatively,~\cref{fig:lighting_qual} displays our method's superior visual results. In some scenes, such as the stone bear shown in the bottom two rows, CAT3D completely fails to reconcile inconsistent input images into any coherent 3D scene. In other cases, CAT3D reconstructs inaccurate ``cloudy'' scene geometry attempting to explain away changes in lighting. In contrast, our method reconciles highly disparate and sparse observations into a consistent 3D scene, allowing the generation of a high-fidelity NeRF with coherent geometry demonstrated in the inset depth maps. We encourage viewing video comparisons in the supplement.

\subsection{Ablations}

In this section, we show the key contributions of our method by ablating important design decisions. Specifically, we demonstrate the importance of using our simulated inconsistency data by evaluating against heuristic augmentations and a synthetic data alternative. For dynamics, we evaluate against a warping-based heuristic augmentation where we apply sparse flow fields; an example can be seen in ~\cref{fig:data}. Interestingly, we find that the resultant model simply copies all ``real-looking'' pixels, indicating that such warping does not adequately bridge the domain gap to real motions. 

We also compare to the alternate approach of generating a synthetic training dataset by animating 40k+ Objaverse assets~\cite{Deitke2022ObjaverseAU, diffusion4d} with associated motions. Due to the small motion magnitude and the domain gap from object-level renderings to real scene-level data, the method significantly underperforms.
The quantitative ablation results can be seen in the top of~\cref{tab:ablations}, where our method outperforms all ablated methodologies. For dynamics, a qualitative comparison is provided on the right of~\cref{fig:dynamics_qual}. 

For lighting, we compare against heuristic augmentation whereby the input images are tinted inconsistently as seen in ~\cref{fig:data}, and the targets images are tinted consistently. This method slightly outperforms the vanilla CAT3D as it requires the model to get the mean color correct; however, it cannot resolve lighting phenomena like shadows, nor localized changes in lighting. Results can be seen quantitatively in~\cref{tab:ablations} and qualitatively on the right of~\cref{fig:lighting_qual}.

\begin{table}[ht]
\centering
\begin{tabular}{@{}c@{\,\,}|l|@{\,}c@{\,}@{\,}c@{\,}@{\,}c@{}}
        &            Ablation           & \,\, PSNR$\uparrow$ & \,\,SSIM$\uparrow$ & \,\,LPIPS$\downarrow$ \\ \midrule 
\multirow{3}{*}{\centering{\rotatebox{90}{\small{Dynamic}}} }
        & Heuristic Augmentation        &         15.52                &            0.448              &     0.466
                        \\
        & Animated Objaverse   &        14.92                 &          0.380                &    0.524                        \\
        & Our Complete Model                  &     \textbf{  16.60      }            &        \textbf{ 0.462         }        &   \textbf{0.409 }                         \\ \midrule %
\multirow{2}{*}{\centering{\rotatebox{90}{\small{Light}}} }
        &    Heuristic Augmentation            &          18.96              &                 0.645         &     0.406                        \\
        & Our Complete Model                  &   \textbf{  20.98      }              &  \textbf{  0.707           }           &      \textbf{      0.357 }                
\end{tabular}

\caption{Ablations. Heuristic augmentation and synthetic datasets lead to significantly worse performance for robust view synthesis. For both inconsistencies in dynamics and lighting, our complete model vastly outperforms the baselines due to the underlying video model's ability to simulate physics. }
\label{tab:ablations}
\end{table}

%% file: 10_conclusion.tex
\section{Discussion}
\label{sec:conclusion}

We have proposed SimVS, a method for high-quality 3D generation from casual captures even in the presence of severe illumination changes and significant scene motion. We believe this represents a step forward in simplifying the capture and creation of 3D scenes. 

\paragraph{Limitations} 
Our method requires accurate camera poses, which can be difficult to compute for sparse captures with significant inconsistencies using traditional techniques such as COLMAP. However, recent methods such as DUST3R~\cite{wang2024dust3r} and the dynamics-robust follow-up MonST3R~\cite{monst3r} have shown tremendous promise for camera pose estimation. When there is very little overlap between views, our method can struggle to reconcile the given observations. 

\paragraph{Conclusion}
Our work demonstrates the power of using video models to generate data for challenging tasks where collection is expensive and challenging. We believe the approach proposed here will scale well with the ever-improving quality of video models. Moreover, our method is not specific to a particular architecture or task: our method may be applied to make DUSt3R~\cite{wang2024dust3r}-style models more robust and our harmonization network could be implemented with a camera-controlled video model to directly synthesize multiview-consistent videos in one sampling pass.

\section*{Acknowledgements}
We would like to thank Paul-Edouard Sarlin, Jiamu Sun, Songyou Peng, Richard Tucker, Linyi Jin, Rick Szeliski and Stan Szymanowicz for insightful conversations and help. We also extend our gratitude to Shlomi Fruchter, Kevin Murphy, Mohammad Babaeizadeh, Han Zhang and Amir Hertz for training the base text-to-image latent diffusion model. This work was supported in part by an NSF Fellowship, ONR grant N00014-23-1-2526, gifts from Google, Adobe, Qualcomm and Rembrand, the Ronald L. Graham Chair, and the UC San Diego Center for Visual Computing.

%% file: 12_appendix.tex
\section{Supplemental Video}
Alongside this PDF, we provide a supplemental video of video results, comparisons to baselines, and ablations. We highly encourage the reader to view the video supplement. 

\section{Evaluation Details}
\label{sec:evaluation}
We use the pretrained CAT3D~\cite{cat3d} model provided by the authors for the lighting benchmarks along with the default implementation of ZipNeRF with GLO~\cite{barron2023zipnerf}. For the dynamics benchmark, we use a CAT3D model finetuned to condition on $5$ images and predict $3$. The lower variance of the conditioning provides a slight benefit as seen in~\cref{tab:supp_ablation}.

\begin{table}[ht]
\centering
\begin{tabular}{@{}c@{\,\,}|l|@{\,}c@{\,}@{\,}c@{\,}@{\,}c@{}}
        & Ablation                  & \,\, PSNR$\uparrow$ & \,\,SSIM$\uparrow$ & \,\,LPIPS$\downarrow$ \\ \midrule 
\multirow{2}{*}{\centering{\rotatebox{90}{}}}
        & w/o $5$ Cond. CAT3D       & 16.57              & 0.453              & 0.414                 \\
        & Our Complete Model        & \textbf{16.60}     & \textbf{0.462}     &           \textbf{0.409 }              \\ 
\end{tabular}
\caption{Performance comparison of ablation conditions.}
\label{tab:supp_ablation}
\end{table}

\section{Comparison to Shape of Motion~\cite{som2024}}
\label{sec:som}
We include an additional baseline comparison to Shape of Motion~\cite{som2024}, the current state-of-the-art method for 4D reconstruction. We consider this type of method to be slightly orthogonal to our approach; incorporating priors such as static masks and monocular depth may improve our results further. 

As in our experiments in the main paper, we provide this baseline with a set of unordered sparse images from the DyCheck~\cite{dycheck} dataset. We compare only on the scenes that Shape of Motion benchmarked, and therefore exclude Space-Out and Wheel. 

\begin{figure}[htbp]
    \centering
    \includegraphics[width=\columnwidth]{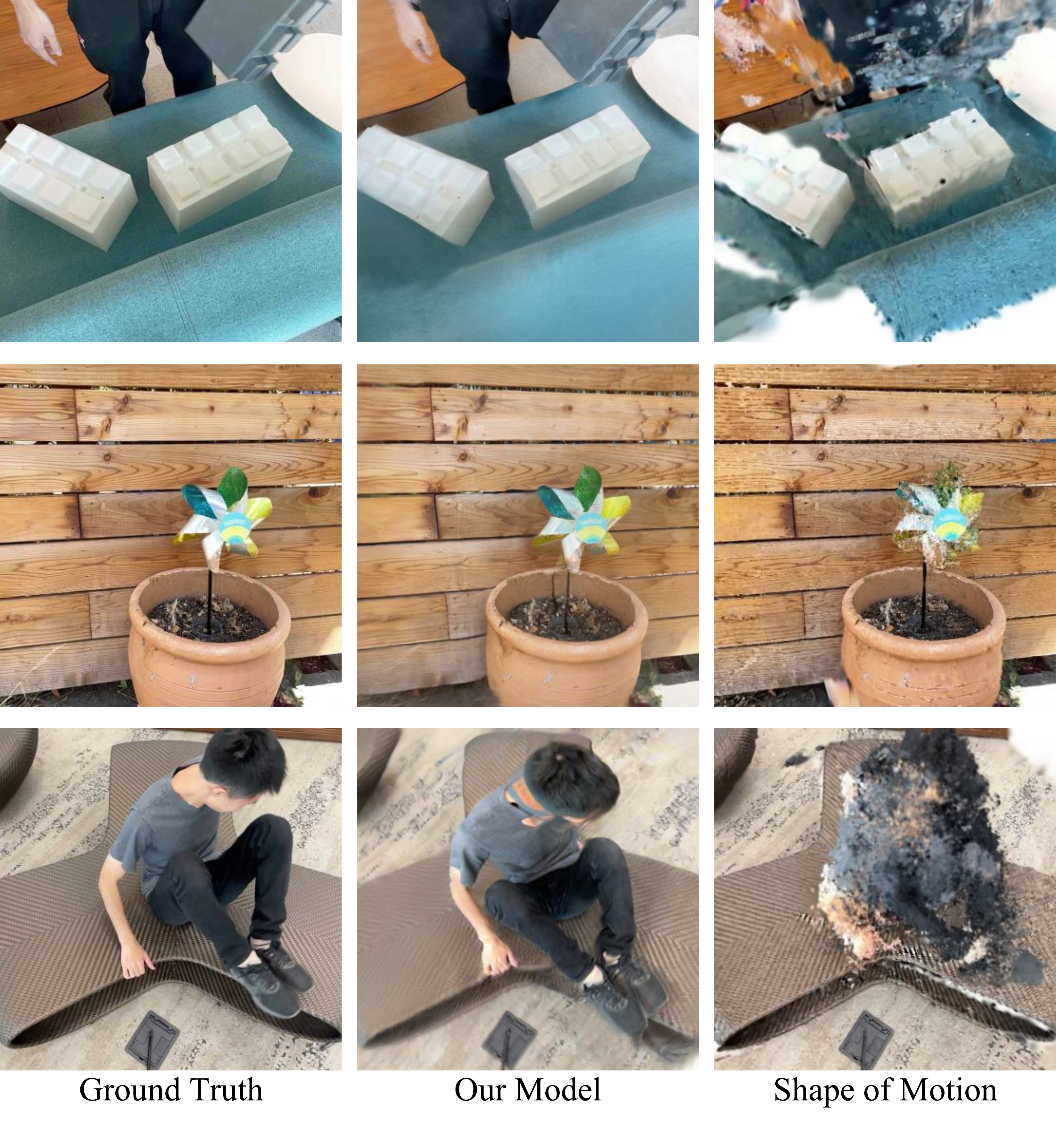} %
    \caption{Qualitative comparison to Shape of Motion~\cite{som2024} on sparse input views from the DyCheck dataset.}
    \label{fig:som}
\end{figure}

We use the refined poses and aligned depth from the original paper and train the model to render the standard $360$x$480$ images, center-cropped to a square aspect ratio as in the comparisons included in the main paper. As specified in their GitHub repository, we computed the video masks with Track Anything~\cite{trackanything}, which shows some robustness to the sparse inputs. However, TAPIR~\cite{tapir} seemed to struggle to compute reasonable tracks given sparse inputs. We show qualitative results in~\cref{fig:som} and quantitative results in~\cref{table:som}. Due to the inability of Shape of Motion to predict scene content outside of the frustums of the input images, we show results with covisibility masks as well. As evidenced by the metrics and qualitative results, Shape of Motion struggles to recover a cohesive representation under the sparse and unordered input setting of this paper. 

\begin{table}[ht]
\centering
\begin{tabular}{@{}c@{\,\,}|l|@{\,}c@{\,}@{\,}c@{\,}@{\,}c@{}}
        & Condition                & \,\, PSNR$\uparrow$ & \,\,SSIM$\uparrow$ & \,\,LPIPS$\downarrow$ \\ \midrule 
\multirow{2}{*}{\centering{\rotatebox{90}{\small{raw}}}}
        & Ours                     & \textbf{16.46}     & \textbf{0.425}     & \textbf{0.484}                \\
        & Shape of Motion~\cite{som2024}                      & 14.10              & 0.396              & 0.485        \\ \midrule
\multirow{2}{*}{\centering{\rotatebox{90}{\small{mask}}}}
        & Ours                     & \textbf{17.03}     & \textbf{0.557}     & 0.410     \\
        & Shape of Motion~\cite{som2024}                      & 15.58              & 0.536              & \textbf{0.391}               \\ 

\end{tabular}
\caption{Performance comparison of methods with and without covisibility masks from~\cite{dycheck}.}
\label{table:som}
\end{table}

\section{Additional visualizations}
\begin{figure*}[htbp]
    \centering
    \includegraphics[width=\textwidth]{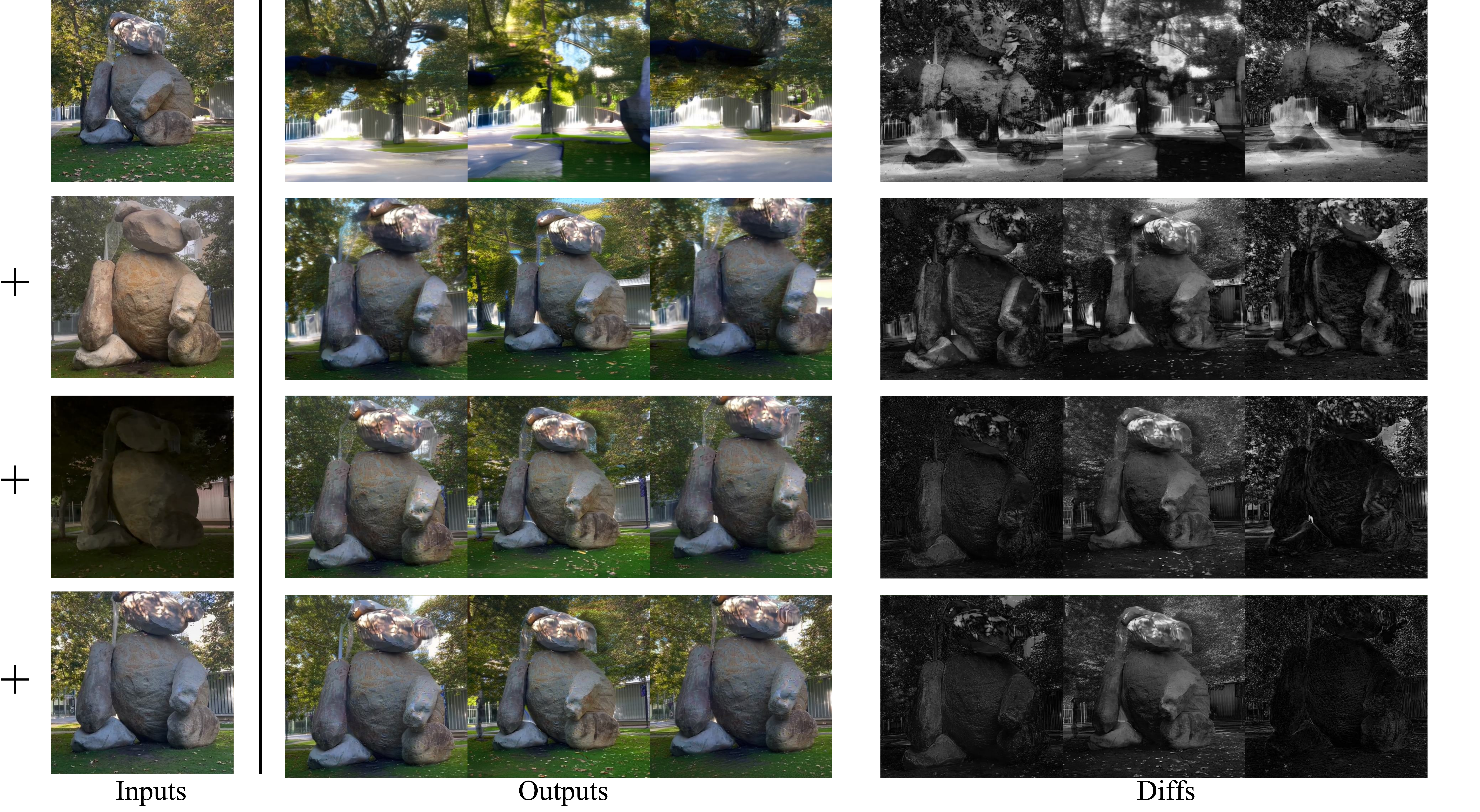} %
    \caption{Our model incorporating more context given an increasing number of images. Given the (additional) inputs on the left, our model reduces uncertainty in its predictions and predicts more well-aligned images to three extra input images as seen in the difference map between additional inputs and the outputs.}
    \label{fig:context}
\end{figure*}

\begin{figure*}[htbp]
    \centering
    \includegraphics[width=\textwidth]{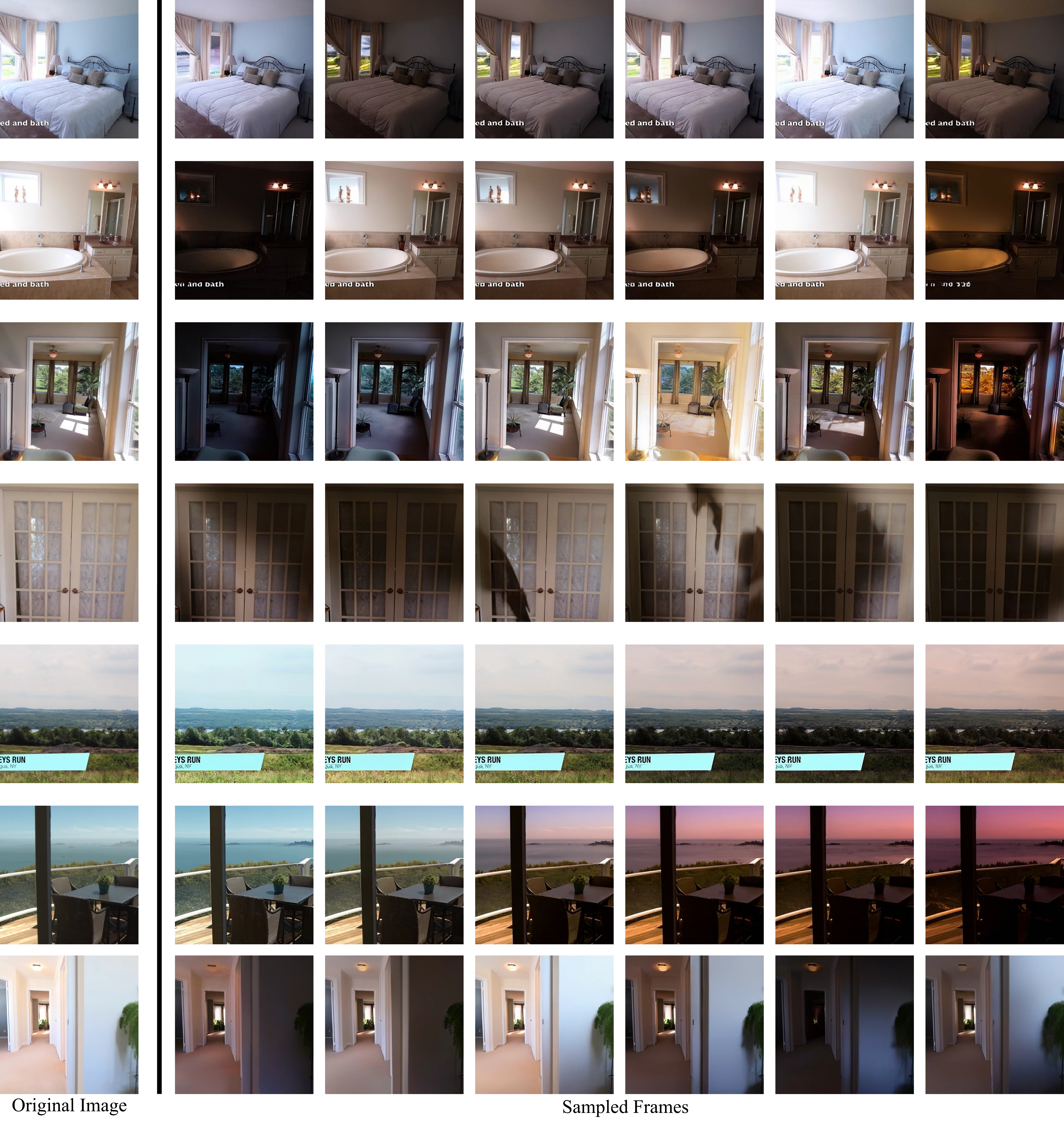} %
    \caption{We show example samples from the lighting data we sampled.}
    \label{fig:samples}
\end{figure*}

We show the ability of our model to effectively and flexibly incorporate more information in~\cref{fig:context}, reducing the uncertainty in its prediction with larger context. We also show samples from the lighting dataset in~\cref{fig:samples}. Due to privacy concerns, we do not show samples from the dynamics dataset, which consists of humans. 

\section{Training Details}
\label{sec:training}
We finetune the pretrained CAT3D~\cite{cat3d} model with $0$ initialization for the input conditioning convolution layer to accept the inconsistent latents $\Tilde{z}$. We train with a batch size of 64 (sets of multiview images) per gradient step. For the lighting model, we finetune for 36k iterations, and for dynamics, 48k iterations. 

We train all ablations for the same amount of time as the corresponding model for the respective data types, except for the dynamics augmentation model which quickly overfits to copying; therefore, we train it for only 12k iterations, as this is where the loss on the held-out OOD data is minimized.

\section{Video Model Prompt Details}
\label{sec:prompt}
In this section, we specify the details of the prompting for the video model including the meta-prompt, example prompts, and list of prompts for lighting. 

\subsection{Lighting prompts}

For lighting, we sample the prompts from the following set: 

\begin{enumerate}
    \item \texttt{"a bright light casts shadows"}
    \item \texttt{"the light slowly dims from bright to dark"}
    \item \texttt{"an object flies around the room, casting hard shadows"}
    \item \texttt{"a transition from a bright day to a dark night"}
    \item \texttt{"the shadows and lights move"}
    \item \texttt{"a strobe light flashes"}
\end{enumerate}

\subsection{Dynamics prompts}

For dynamics, we sample about 10k total prompts using the meta-prompt given in~\cref{fig:meta_prompt}. We include $20$ examples below:

\begin{enumerate}[leftmargin=*]
    \item \texttt{"They walk quickly along the path, the child struggling to keep up while carrying the bottle."}
    \item \texttt{"The boys playfully pose for a photo."}
    \item \texttt{"The mechanics are actively repairing the car, with tools moving and parts being replaced."}
    \item \texttt{"The girls are collaboratively typing on the laptop."}
    \item \texttt{"The chef moves through the train serving food to passengers."}
    \item \texttt{"Children run through the play tunnel and climb onto the boat."}
    \item \texttt{"The children run around the line, crossing it repeatedly during the game."}
    \item \texttt{"The girl walks past a classroom art display."}
    \item \texttt{"Two people actively select books and papers from the table."}
    \item \texttt{"The puppeteer manipulates the puppets, making them move and interact."}
    \item \texttt{"The woman excitedly raises and lowers her arms."}
    \item \texttt{"The woman gestures emphatically as the man adjusts a component on the truck door."}
    \item \texttt{"The two assistants helped Santa adjust his position in the chair."}
    \item \texttt{"The children reach for items on the table, some stand up and move to a different seat."}
    \item \texttt{"The man gestures emphatically while speaking on the phone."}
    \item \texttt{"The majorette tosses and catches the baton."}
    \item \texttt{"The woman raises and lowers her mug as she drinks."}
    \item \texttt{"The child reached for a cleaning supply."}
    \item \texttt{"The woman dramatically throws her arms out in a wide arc."}
    \item \texttt{"Someone rolled up the red fabric and placed it against the shelf."}
\end{enumerate}

\section{Details of Lumiere sampling}
For sampling from the Lumiere model, we utilize a random-frame variant where the input frame can be anywhere in the video (not just the first frame). This variant is trained by sampling a random frame for each training video and concatenating the input to every frame along the channel dimension, identically as the Lumiere inpainting model. 

We use the following camera-based negative prompt to induce the desired characteristics in the output video and alleviate Lumiere's tendency to output still videos: 
\begin{align*}
    c_\text{negative} &= \texttt{"frozen, photograph, fixed} \\
                      &\quad \texttt{lighting, moving camera, zoom in,} \\
                      &\quad \texttt{zoom out, bird view, panning view,} \\
                      &\quad \texttt{360-degree shot, orbit shot,} \\
                      &\quad \texttt{arch shot"}
\end{align*}
We use $250$ DDPM sampling steps for the image- and text-conditioned Lumiere base model at a resolution of $128$x$128$. We then upsample that video conditioned only the original prompt to a size of $1024$x$1024$ with $250$ sampling steps and resize to the desired size of $512$x$512$. We set the guidance weight to $6$ for both processes.

\begin{figure*}[htbp]
    \centering
    \includegraphics[width=\textwidth]{figs/meta_prompt.pdf} %
    \caption{The meta-prompt used to generate dynamics captions on the Mannequin Challenge dataset~\cite{mannequinchallenge}.}
    \label{fig:meta_prompt}
\end{figure*}

